\documentclass[conference]{IEEEtran}

\IEEEoverridecommandlockouts                      

\usepackage{graphics} 
\usepackage{epsfig} 
\usepackage{mathptmx} 
\usepackage{times} 
\usepackage{amsmath} 
\usepackage{mathtools}
\usepackage{amssymb}  
\usepackage{subcaption}
\usepackage{algorithm}
\usepackage{algpseudocode}
\usepackage{xcolor}
\usepackage{comment}
\usepackage{multirow}
\usepackage{lipsum}
\usepackage{url}
\usepackage{soul}

\usepackage{glossaries}
\algrenewcommand\algorithmicrequire{\textbf{Input:}}
\algrenewcommand\algorithmicensure{\textbf{Output:}}
\DeclarePairedDelimiter\abs{\lvert}{\rvert}%
\DeclarePairedDelimiter\norm{\lVert}{\rVert}%

\makeatletter
\let\oldabs\abs
\def\abs{\@ifstar{\oldabs}{\oldabs*}}

\let\oldnorm\norm
\def\norm{\@ifstar{\oldnorm}{\oldnorm*}}
\makeatother

\newcommand{\h}{\mathbf{h}}

\renewcommand{\H}{\mathbf{H}}

\newcommand{\setA}{\mathcal{A}}

\newcommand{\setC}{\mathcal{C}}
\newcommand{\setD}{\mathcal{D}}
\newcommand{\setE}{\mathcal{E}}

\newcommand{\setI}{\mathcal{I}}

\newcommand{\setP}{\mathcal{P}}

\newcommand{\setR}{\mathcal{R}}

\newcommand{\Real}{\mbox{$\mathbb{R}$}}

\newacronym{isac}{ISAC}{integrated sensing and communication}
\newacronym{lidar}{LiDAR}{light detection and ranging}
\newacronym{gps}{GPS}{global positioning system}
\newacronym{ai}{AI}{artificial intelligence}
\newacronym{em}{EM}{electro magnetic}
\newacronym{csi}{CSI}{channel state information}
\newacronym{los}{LoS}{line-of-sight}
\newacronym{nlos}{nLoS}{non-line-of-sight}
\newacronym{6g}{6G}{sixth-generation}
\newacronym{hdc}{HDC}{hyperdimensional computing}
\newacronym{hv}{HV}{hypervector}
\newacronym{cir}{CIR}{channel impulse response}
\newacronym{map}{MAP}{multiply-add-permute}
\newacronym{xor}{XOR}{exclusive-OR}
\newacronym{agv}{AGV}{automated guided vehicle}
\newacronym{cots}{COTS}{commercial off-the-shelf}
\newacronym{fov}{FoV}{field-of-view}
\newacronym{mcu}{MCU}{microcontroller unit}
\newacronym{cs}{CS}{cosine similarity}
\newacronym{hd}{HD}{Hamming distance}
\newacronym{mimo}{MIMO}{multiple-input multiple-output}
\newacronym{rx}{RX}{receiver}
\newacronym{tx}{TX}{transmitter}
\newacronym{hdm}{HD}{hyperdimensional}
\newacronym{az}{AZ}{azimuth}
\newacronym{el}{EL}{elevation}
\newacronym{dnn}{DNN}{deep neural network}
\newacronym{fl}{FL}{forklift}
\newacronym{ros}{ROS}{robot operating system}

\newcommand{\CIR}{\mathrm{CIR}}

\newcommand{\pwr}{\scriptscriptstyle \mathrm{PWR}}
\newcommand{\del}{\scriptscriptstyle \mathrm{DEL}}
\newcommand{\az}{\scriptscriptstyle \mathrm{AZ}}
\newcommand{\el}{\scriptscriptstyle \mathrm{EL}}
\newcommand{\RX}{\scriptscriptstyle \mathrm{RX}}

\title{
Leveraging Distributed Sensing and Hyperdimensional Computing for Non-Line-of-Sight Passive Object Sensing
}

\author{Placido Mursia, Marco Rossanese, Francesco Devoti, Adrián Lendínez Ibáñez,\\ Lanfranco Zanzi, Xi Li, Xavier Costa-P\'erez
\thanks{This work has been partially funded by the SNS JU projects 6G-DISAC, INSTINCT, and Multi-X (Grant Agreement No 101139130, 101139161, and 101192521, respectively). P. Mursia, M. Rossanese, F. Devoti, L. Zanzi, Xi Li, are with NEC Laboratories Europe GmbH, Heidelberg, Germany. {\tt\small \{name.surname\}@neclab.eu}. A. Lendínez Ibáñez is with Univeristy of Bedfordshire, Luton, United Kingdom. {\tt\small adrian.lendinezibanez@study.beds.ac.uk}. X. Costa-Pérez is with NEC Laboratories Europe GmbH, Heidelberg, Germany, i2CAT Foundation, Barcelona, Spain, and ICREA Foundation, Barcelona, Spain. {\tt\small xavier.costa@ieee.org}}%
}

\begin{document}

\maketitle

\begin{abstract}
In smart factory environments, passive entities such as humans and movable obstacles may interfere with autonomous robot operations, posing significant safety risks. Detecting potential conflicts between robots and entities outside their \gls{fov} is therefore of paramount importance. 
However, conventional passive object detection systems rely on sensors such as LiDAR, radar, or cameras, which operate under direct \gls{los} and are limited by obstacles and occlusions.
To overcome these limitations, \gls{isac} enables environmental perception via the communication infrastructure thanks to cooperation among distributed wireless devices. 
In this paper, we propose a 
framework for passive entity sensing beyond the robot's \gls{fov} that relies solely on distributed channel measurements from the communication infrastructure to perform 
risk-aware classification. 
These measurements are embedded into high-dimensional representations using \gls{hdc}, enabling robust and scalable learning without explicit channel reconstruction. Numerical simulations based on a commercial ray tracer and ROS-based validation in realistic smart factory scenarios demonstrate improved safety and enhanced path planning with up to 89$\%$ accuracy under severe occlusions, while maintaining low computational complexity. 

\end{abstract}

\glsresetall
\section{Introduction}

\Gls{isac} represents a paradigm shift in \acrshort{6g} wireless systems, inherently coupling communication and sensing~\cite{strinati2024distributed}. It promises unprecedented gains in throughput and latency while integrating \gls{ai}, 3D coverage, and turning the wireless infrastructure into a distributed environmental perception layer~\cite{droulias2026integrating}.

This is particularly relevant in dynamic industrial scenarios, where humans and autonomous robots move through shared spaces and risk collisions, injuries, and operational disruptions~\cite{maresca2026localization}. In such environments, detecting passive entities, such as humans, nomadic or stationary objects that are not connected to sensing or communication systems, is therefore crucial for safety and reliability. Traditional onboard sensing based solely on LiDAR or cameras on the robot is constrained by the direct \gls{los} \gls{fov} of the sensor, and further limited by occlusions, walls, sensing range, and computational capacity. In this context, \gls{isac} allows robots to leverage the mobile network as an external perception layer~\cite{jia2024infrastructure}, obtaining contextual information about their surroundings--including areas beyond their direct \gls{fov}--and enabling network-aided navigation and collision avoidance. 

Despite recent advances enabled by 5G systems and higher carrier frequencies, sensing passive entities remains challenging. Radar-like methods process backscattered echoes for object detection and human motion tracking~\cite{Pegoraro24}, but often require complex hardware and signal processing~\cite{Zhao18}. Lower-complexity alternatives exploit power measurements~\cite{Devoti20}, received signal strength~\cite{sun2025revealing}, or channel fluctuations~\cite{yang2013rssi} to infer environmental changes from communication signals.

\begin{figure}[t!]
    \centering
    \includegraphics[width=\linewidth]{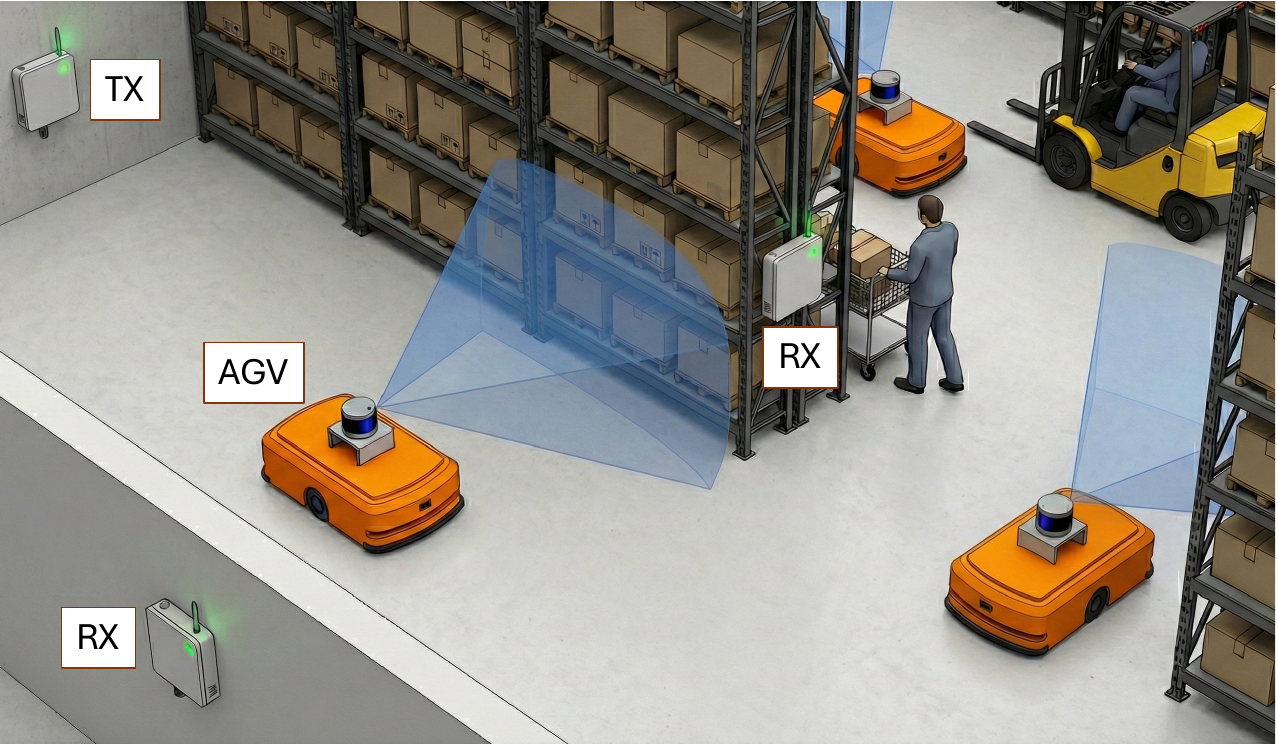}
    \caption{Considered smart factory scenario where humans and robots operate concurrently.}
    \label{fig:sys-model}
    \vspace{-3mm}
\end{figure}

Unlike existing \gls{isac} approaches relying on static sensing nodes and operating mainly under \gls{los} conditions, we propose cooperative passive sensing 
of regions outside the robot's direct sensing visibility. Since passive objects located around corners or behind shelving structures may not generate observable signatures within the robot's \gls{fov}, the proposed framework leverages spatially distributed 
channel measurements, thereby complementing conventional sensing modalities. We consider an indoor smart factory where humans and robots operate concurrently, and obstacles may be just behind or around corners, which are invisible to a robot’s onboard sensors, as depicted in Fig.~\ref{fig:sys-model}. 
By leveraging the distributed wireless infrastructure as a collaborative perception layer, our approach infers environmental changes 
and provides preemptive information on unseen passive objects, which might impact its planned trajectory, enabling proactive route adaptation and enhanced safety.
\begin{figure*}[t!]
    \centering
    \includegraphics[width=0.9\linewidth]{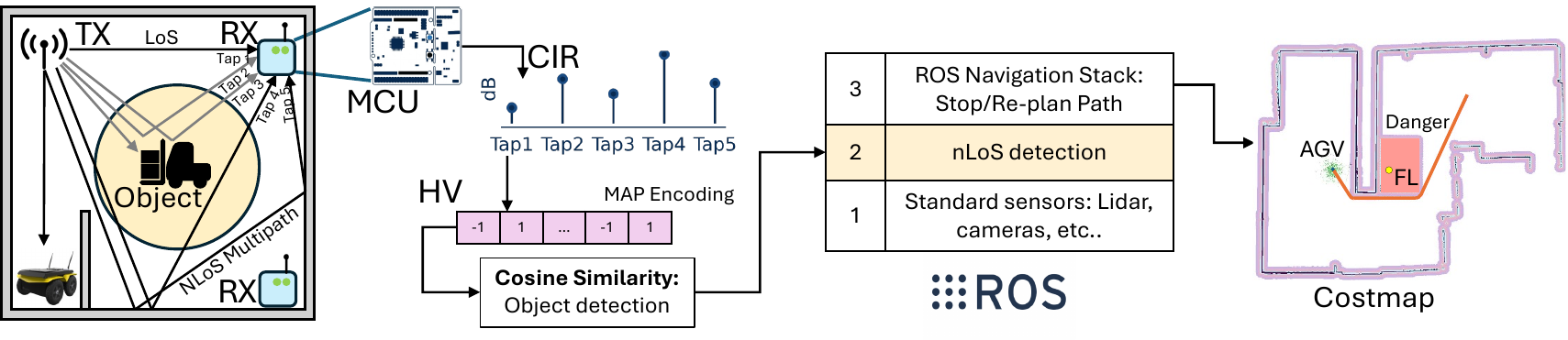}
    \caption{Proposed system architecture combining distributed wireless sensing and \acrshort{hdc} to complement onboard robot sensors, allowing them to preemptively perform risk assessment of passive entities beyond their \acrshort{fov}, and update their path.}
    \label{fig:architecture}
    \vspace{-3mm}
\end{figure*}
To achieve this, we combine the spatial diversity obtained through coordinated distributed sensing with the \gls{hdc} paradigm, supporting efficient learning in complex environments through high-dimensional representations~\cite{Lulu20}. While \gls{hdc} has shown promise in several domains (see, e.g.,~\cite{lee2025hypercam, Eman22}), its use in wireless sensing remains largely unexplored. 
This work therefore provides four main contributions: \textbf{c1)} a cooperative infrastructure-assisted distributed \gls{isac} framework for risk-aware sensing of passive entities beyond the robot sensing range, \textbf{c2)} a low-complexity \gls{nlos} sensing method based on channel measurements, \textbf{c3)} a robust and scalable \gls{hdc}–based sensing scheme, and \textbf{c4)} validation through extensive ray-tracing and \gls{ros}-based experiments, and running on typical robot \gls{mcu} hardware.


\section{System model}\label{sec:sys_model}

We consider an indoor smart factory setup, as depicted in Fig.~\ref{fig:sys-model}, where robots assist humans via conventional mapping tools (e.g., LiDAR), and follow pre-determined and/or dynamically adjustable trajectories. We assume that a single-antenna \gls{tx} is deployed alongside multiple multi-antenna \glspl{rx}. Due to the complex nature of such a propagation environment,
an \gls{agv} may have a limited \gls{fov}, i.e., its \gls{los} may be obstructed, preventing it from discovering all obstacles using conventional techniques.
For this reason, we divide the factory floor into two areas, namely \textit{safe} and \textit{danger} areas, where the latter represents the obstructed \gls{fov} region where collisions with humans might occur.
We assume that \glspl{rx} enable precise beamforming by steering the main beam towards the direction $(\theta,\phi)$ in \gls{az} and \gls{el}, respectively.
Under strict synchronization, the \gls{tx} emits a known reference signal, while each \gls{rx} extracts the \gls{cir} for each beamforming direction in a multistatic fashion. We consider a passive target (i.e., not networked) located at $(x_k, y_k, 0)$ on the factory floor, at the $k$-th time slot.

The overall system architecture is shown in Fig.~\ref{fig:architecture}. The passive object, located outside the \gls{fov} of the robot, is sensed through \glspl{cir} collected at multiple \glspl{rx}. These are then encoded into \glspl{hv} and used to infer the object class (i.e., \emph{safe} or \emph{danger}), as detailed in the following sections.
We formulate the detection task as a binary classification problem, since it serves as a foundational proof-of-concept to validate the feasibility of \gls{hdc}-based \gls{nlos} sensing before extending to high-resolution sensing tasks such as localization, trajectory estimation, and multi-class scene understanding, which are left for future work. If the target is within the danger zone, the system provides preemptive \gls{nlos} information to enhance proactive environmental awareness. The \gls{ros} navigation stack then uses this output to adapt the robot's path. The proposed method is therefore not intended as a standalone safety-critical system, but as an auxiliary sensing layer that complements existing sensors.

\section{A primer on hyperdimensional computing}
\label{subsec:HDC_primer}

\Gls{hdc} is a brain-inspired computing paradigm for robust and efficient information processing.
It represents data as high-dimensional \glspl{hv}, which enable expressive yet noise-tolerant representations.
\gls{hv}s are commonly binary or integer-valued with independent and identically distributed (i.i.d.) components, resulting in distributed representations where information is spread across all dimensions~$D$, typically in the thousands or tens of thousands.
Randomly generated \gls{hv}s are nearly orthogonal to one another, which supports reliable similarity-based retrieval and computation in superposition. Algebraic operations such as \textit{binding, bundling, and permutation} maintain the same dimensionality, thus improving scalability and energy efficiency compared to tensor product representations such as neural network-based approaches~\cite{Lulu20}.


\noindent \textbf{\underline{\textit{Similarity Measurement}}.} Two methods are used to compare \gls{hv}s: \gls{cs} for non-binary and \gls{hd} for binary vectors. 
The \gls{cs} is determined by the angle between vectors rather than their magnitude: $\cos(\mathbf{A}, \mathbf{B}) = (\mathbf{A} \cdot \mathbf{B})/(\|\mathbf{A}\| \|\mathbf{B}\|)$.
For binary \gls{hv}s with dimensionality $D$, 
the normalized \gls{hd} is 
    $\text{Ham}(\mathbf{A}, \mathbf{B}) = \frac{1}{D} \sum_{i=1}^{D} |A_i - B_i|$.
For \gls{cs}, a value of $\{1,\,0\,,-1\}$ represents identical, orthogonal, and diametrically opposed vectors, respectively, while the same applies for values $\{0,\,0.5,\,1\}$ of \gls{hd}, respectively.


\noindent \textbf{\underline{\textit{Bundling}}.} Pointwise addition, denoted as $\oplus$, aggregates a set of input \gls{hv}s $\{\mathbf{X}_1, \mathbf{X}_2, \dots, \mathbf{X}_n\}$ into a single composite \gls{hv} $\mathbf{Z}$, 
which is maximally similar to any of the $n$ vectors.
The bundling operation strictly follows the addition rules defined by the specific working space of the \gls{hv}s. For binary \gls{hv}s, the summation output is binarized via the majority rule~\cite{Lulu20}.


\noindent \textbf{\underline{\textit{Binding}}.}
The pointwise multiplication operation, denoted as $\otimes$, forms associations between two related \gls{hv}s, $\mathbf{A}$ and $\mathbf{B}$. For binary \gls{hv}s, binding is implemented via component-wise \gls{xor} operation, while for the non-binary case it follows the multiplication rules defined by their specific working space. Unlike bundling, binding produces a result that is approximately orthogonal to both operands.


\noindent \textbf{\underline{\textit{Permutation}}.} Permutation, denoted by $\pi(i)$, circularly shifts the components of a single \gls{hv} by $i$ positions. The resulting \gls{hv} is quasi-orthogonal to the original one, enabling differentiation between identical elements appearing in different orders (e.g., distinguishing “A then B” from vice versa).

\section{Hyperdimensional computing framework}\label{sec:HDC}

We consider a dataset of \gls{cir} measurements collected at multiple \glspl{rx} under different beamforming configurations and target positions. Each sample corresponds to a \gls{cir} observation associated with a given \gls{rx}, beam direction, and time slot. The goal of the proposed framework is to encode such measurements into \glspl{hv} for classification. In particular,
we employ the \gls{map} encoding, which is represented by a mapping function $\mathcal{H}: \Real \to \{-1,+1\}^D$ converting a number $x\in\Real$ into a uniform binary vector as
\begin{align}\label{eq:H_ref}
    \mathcal{H}(x) = \{h_k\}_{l=1}^D, \quad h_l\sim \mathrm{U}[-1, +1] \quad \forall l.
\end{align}

Throughout the paper, we distinguish between continuous and binary \gls{map} encodings. In the continuous case, intermediate \gls{hv} representations preserve real-valued accumulation results and similarity is computed using \gls{cs}. In the binary version, the accumulated \gls{hv} is subsequently binarized through the sign operator, enabling lower-complexity storage and processing.
Given the beam sweeping range at the \glspl{rx}, $\theta \in [\theta_{\min},\theta_{\max}]$ and $\phi \in [\phi_{\min},\phi_{\max}]$ in the \gls{az} and \gls{el} directions, respectively, and given a discrete quantization of both ranges $\setA$ and $\setE$, respectively. We construct the dictionaries $H_{\az}$ and $H_{\el}$ as
\begin{align}
    H_{\az} = \{\mathcal{H}(\theta),\,\, \forall \theta \in \setA\},\quad 
    H_{\el} = \{\mathcal{H}(\phi),\,\, \forall \phi \in \setE\}.
\end{align}
Similarly, given a range of delays $[0,\tau_{\max}]$, powers $[0,p_{\max}]$, RX IDs, and associated discretized sets $\setD$, $\setP$, and $\setR$, we construct the dictionaries $H_{\del}$, $H_{\pwr}$, and $H_{\RX}$, respectively.

\begin{figure}[t!]
    \centering
    \includegraphics[width=0.8\linewidth]{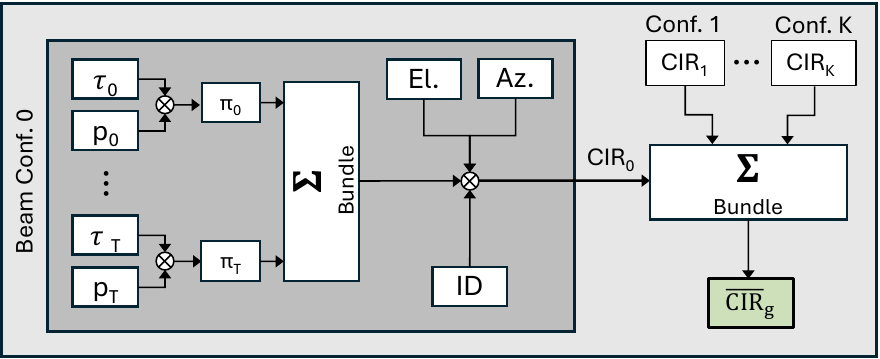}
    \caption{Group-level encoding for a given target position.}
    \label{fig:group_enc}
    \vspace{-3mm}
\end{figure}

\subsection{Group-level encoding}\label{subsec:group_enc}

Each dataset row represents a \gls{cir} measured at a given \gls{rx} with a fixed beam configuration and target position at time slot $k$, labelled as safe or danger. Since, for each target position, every \gls{rx} records the \gls{cir} across all beam configurations, we divide the dataset into groups, where each group contains all measurements from all \glspl{rx} for a fixed target position.

For each row in a group, the \gls{cir} consists of $T$ \textit{taps}, i.e., pairs of delays and power values representing multipath components including \gls{los} and \gls{nlos} paths. We bind the information of each tap into a single \gls{hv}. Moreover, since the order in which each channel tap is received matters, we permute the resulting \gls{hv} by $i$, where $i$ is the tap number. The resulting \gls{cir} signature is bound with the \glspl{hv} associated with the current beam pointing direction (\gls{az} and \gls{el}), and the \gls{rx} ID. The proposed row-level encoding is formalized as
\begin{align}\label{eq:CIR_k}
    \CIR_k =  \left[\sum_{i=1}^{T}\Pi_{i}\left(\h_{\pwr}^{(k,i)}\otimes\h_{\del}^{(k,i)} \right)\right]\otimes\h_{\az}^{(k)}\otimes\h_{\el}^{(k)}\otimes\h_{\RX}^{(k)},
\end{align}
which represents the effective signature of the \gls{cir} at a specific \gls{rx} and associated 3D beamforming configuration, where $k\in\mathcal{I}_g$, and $\setI_g$ is the set of (row) indexes for a given target position, which spans the full 3D beamsweep. Finally,
\begin{align}
    \CIR_g = \sum_{k\in\mathcal{I}_g}\CIR_k,
\end{align}
represents the overall channel signature for a fixed target location. Fig.~\ref{fig:group_enc} summarizes the group-level encoding process.

\begin{figure}[t!]
    \centering
    \includegraphics[width=0.8\linewidth]{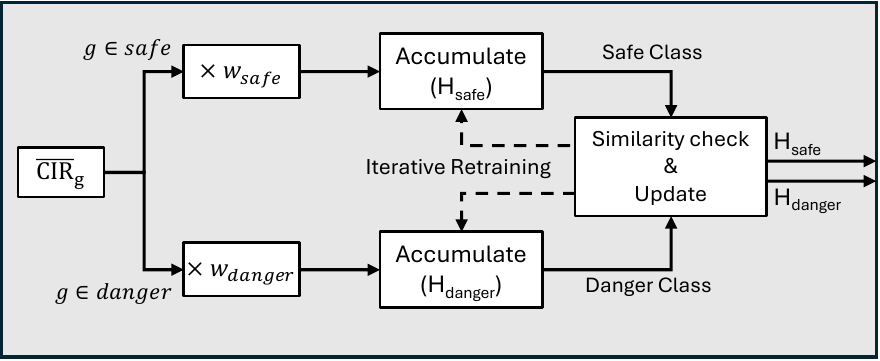}
    \caption{Class accumulation with iterative retraining tuning.}
    \label{fig:class_acc}
    \vspace{-3mm}
\end{figure}

\subsection{Training}\label{subsec:training}

Our goal is to build a classifier that can infer whether the target is in the danger zone or not. 
To this end, we construct a single \gls{hv} per class (\textit{safe}, \textit{danger}) by accumulating the \gls{hdc} representation of their samples.
Since the set of samples of the two classes might differ substantially in size, depending on the operational scenario (in general, \textit{danger} and \textit{safe} areas are not equally sized), we weigh each group level \glspl{hv} $\CIR_g$ by the inverse of the class count. Let $\setC = \{\text{\textit{safe}},\text{\textit{danger}}\}$ be the set of classes, and $n_c$ denote the number of samples in class $c$. The average and normalized class weights are
\begin{align}
    \bar{w} = \frac{1}{|\setC|}\sum_{c\in\setC}\frac{1}{n_c}, \,\,\text{and}\,\,\bar{w}_c =\frac{1}{\bar{w}\,n_c}, \, \forall c \in \setC,
\end{align}
respectively.
Let the labelling function $\mathcal{L}:\,\mathcal{G} \to \mathcal{C}$ map each group to its class, such that the former is weighted as 
\begin{align}
    \overline{\CIR}_g = \bar{w}_{\mathcal{L}(g)} \CIR_g.
\end{align}

At this stage, the \glspl{hv} for each group, $\overline{\CIR}_g$, are bundled with their corresponding class-representative \gls{hv}, $\H_c$, based on their label, i.e., whether the (known) target position belongs to the safe or danger areas. The final
resulting vectors $\{\H_{c}\}_{c\in \mathcal{C}}$ are normalized 
to have entries in $\pm 1$:
\begin{align}
    \H_{\mathcal{L}(g)} = \mathrm{sign}\bigg(\sum_g \overline{\CIR}_g\bigg).
\end{align}

\subsection{Iterative Retraining}
\label{sec:retraining}

While single-pass training via bundling is highly efficient, it may yield overlapping classes in complex feature spaces. To enhance separability, we adopt an iterative retraining strategy inspired by~\cite{AdaptHD},
where the training data is initially shuffled to avoid order bias. Next, for each sample $\overline{\CIR}_g$ with true label $L_g$ at iteration $t$ within an epoch, we compute its cosine similarity with the current class prototypes to obtain the estimated label $\hat{L}_g$. 
Let $\mathbf{1}[\cdot]$ denote the indicator function. If the sample is misclassified, the \gls{hv} prototypes are updated as
\begin{align}
    \H_c^{(t+1)} \!=\! \H_c^{(t)}\! + \!\alpha (\mathbf{1}[c \!=\! L_g] \!-\! \mathbf{1}[c \!=\! \hat{L}_g]), \, \forall c \in \mathcal{C}, \, g\in \mathcal{G},
\end{align}
where $\alpha$ is the learning rate. To preserve the hyperspace properties, prototypes are renormalized at the end of each epoch. The algorithm runs for a given maximum number of epochs, tracking the global minimum of training errors, and performing early stopping upon convergence (zero errors), thus retaining the optimal separation boundary found during the iterative process. This process is depicted in Fig.~\ref{fig:class_acc}. Thanks to this highly computationally efficient procedure, the system can rapidly adapt to environmental changes.

\begin{figure}[t!]
     \centering
     \begin{subfigure}[b]{0.48\linewidth}
         \centering
         \includegraphics[width=\linewidth]{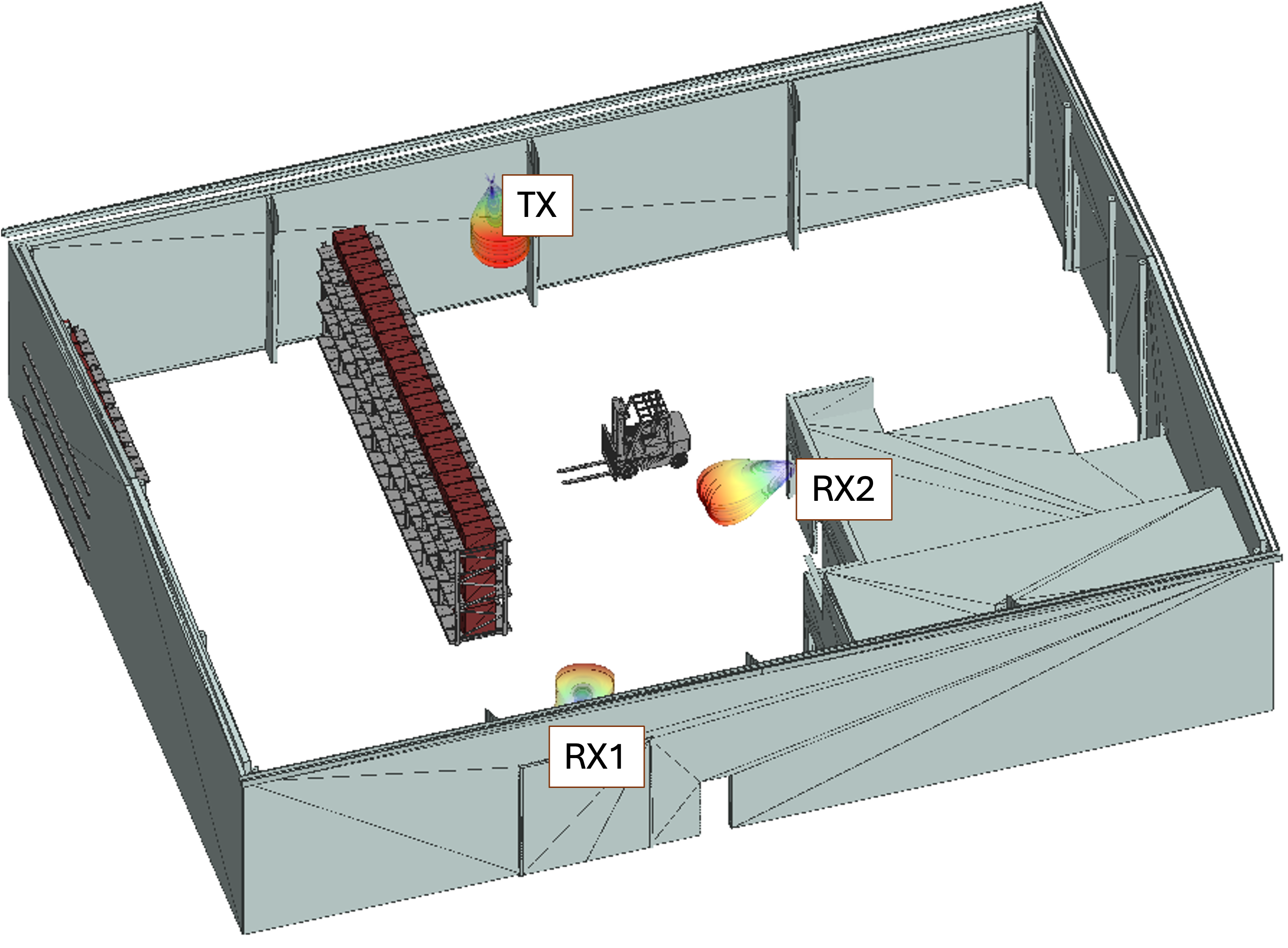}
         \caption{Warehouse environment in the ray-tracer.}
         \label{fig:simul_env}
     \end{subfigure}
     \hfill 
     \begin{subfigure}[b]{0.48\linewidth}
         \centering
         \includegraphics[width=\linewidth]{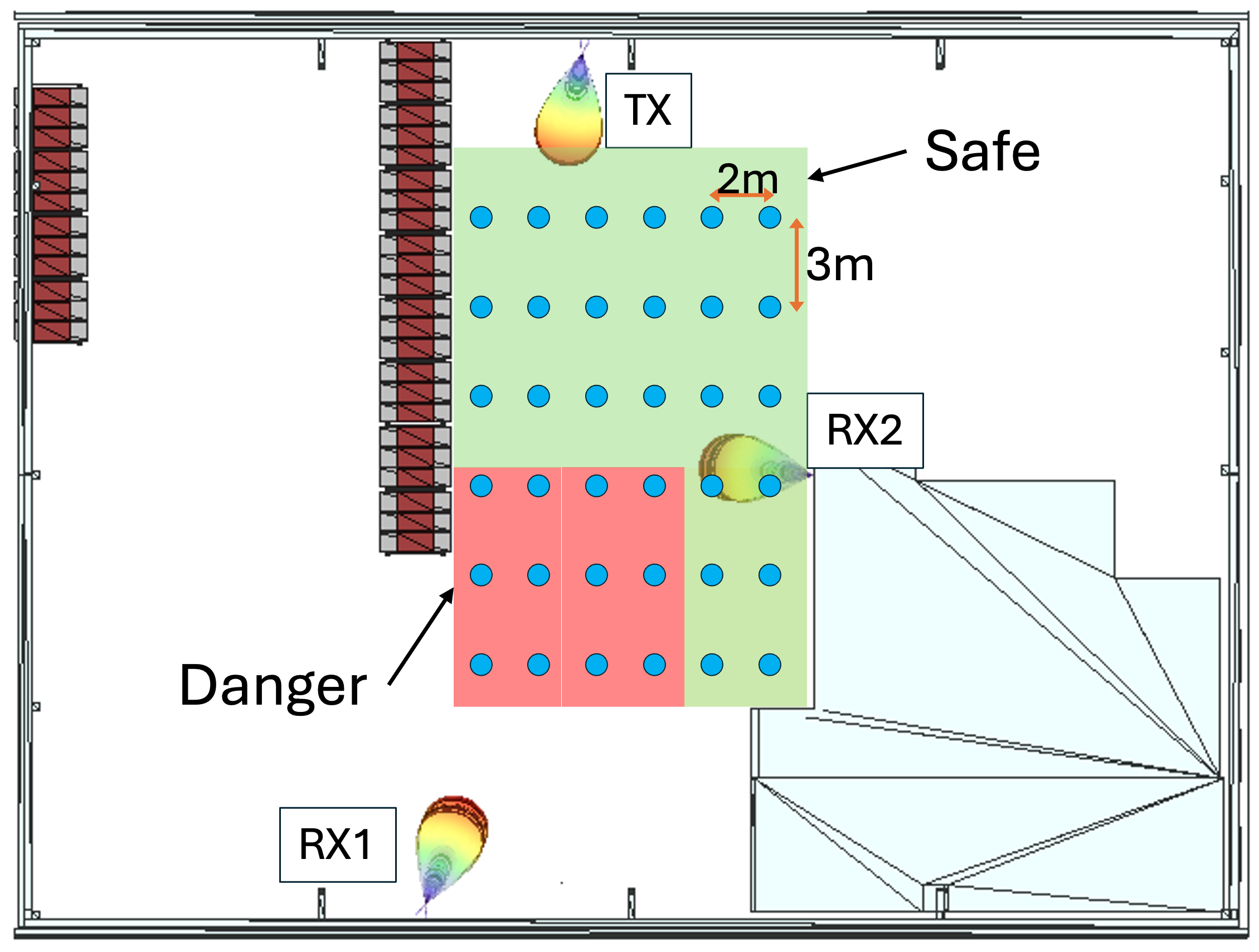}
         \caption{Blue dots: RX beams; Red/Green: Danger/Safe zones.}
         \label{fig:simul_env-top}
     \end{subfigure}
     
     \caption{Simulated warehouse environment layouts.}
     \label{fig:overall_simulation}
     \vspace{-3mm}
\end{figure}

\subsection{Inference}\label{subsec:inference}

Given a set of \glspl{cir} measured across all \glspl{rx} and beamforming configurations, we infer the class $\hat{L}$ of the current unknown target location as follows: all \glspl{rx} perform 3D beamsweeping concurrently, while the \gls{tx} emits known reference signals. The \glspl{cir} from all \glspl{rx} are aggregated into a single fusion module, which performs group encoding, as described in Section~\ref{subsec:group_enc}, thereby producing a single \gls{hv}, $\overline{CIR}_i$. We then compute the similarity between the latter and the representative \gls{hv} of each class $\{\H_{c}\}_{c\in \mathcal{C}}$, where the highest value determines $\hat{L}$.


\section{Numerical results and discussion}\label{sec:NR}

We set up a warehouse scenario depicted in Fig.~\ref{fig:simul_env}. It consists of a moving \gls{fl}, i.e., the passive target, a single-antenna \gls{tx} illuminating the factory floor, and two multi-antenna \glspl{rx}. The former has a Half-Power Beam-Width of $45^\circ$ and emits $20$ dBm, while for the latter it is $15^\circ$; all operate at $30$ GHz.
The \glspl{rx} scan their 3D receive beams towards a grid of 2D discrete locations on the factory floor, depicted with blue dots. The corresponding angles in $H_{\az}$ and $H_{\el}$ are determined by the pointing direction from the \glspl{rx} towards each grid location and are therefore, in general, non-uniformly spaced.
The working area is divided into danger and safe zones, indicated in red and green, respectively. Radio propagation simulations are performed with Wireless InSite, a commercial Ray-Tracer. Rays are cast with $0.3^\circ$ separation, $4$ reflection levels, and $1$ refraction level, with up to $250$ taps for each \gls{cir}. For each \gls{fl} position, a full receive-beam sweep is performed at both \glspl{rx}, and the corresponding \glspl{cir} are collected in a dataset for training (see Section~\ref{subsec:group_enc}). A total of $10^4$ \gls{cir} samples are collected and partitioned into an $80/20$ split for training and testing, following the retraining procedure ($\alpha=0.1$) in Section~\ref{sec:HDC}. We assume the generated \glspl{cir} to be perfectly estimated, thereby focusing on the sensing performance of the proposed \gls{hdc} framework and leaving the impact of imperfect \gls{cir} estimation for future work. The ground truth labels for the training dataset were assigned geographically, based on the target coordinates falling within the safe or danger regions. We set the \gls{hv} dimension to $D=10000$, the maximum retraining epochs to $100$, and average results over $N=1000$ runs to account for the inherent randomness of \gls{hdc} frameworks.

In our experiments, the power distribution of the \glspl{cir} is heavily skewed toward lower values. Hence, we adopt a quantile strategy with $109$ bins ensuring equal data distribution and maximum entropy. Conversely, since the channel tap delays reveal a multimodal distribution, we select a uniform strategy with $267$ bins, maintaining a constant nanoseconds-per-bin ratio and reflecting physical proximity and temporal scale in \gls{hv} similarity.


\begin{figure}[t!]
    \centering
    \includegraphics[width=\linewidth]{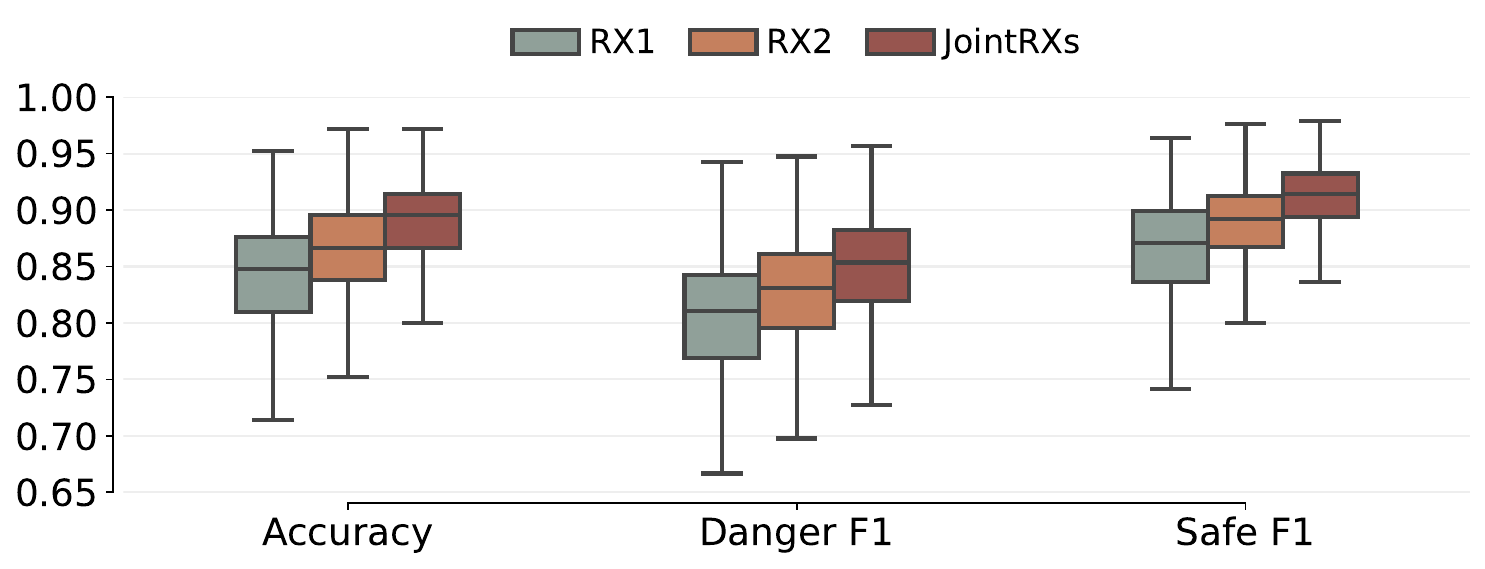}
    \caption{Performance 
    comparison for RX 1, RX 2, and Joint RXs configurations.}
    \label{fig:overall_results}
    \vspace{-3mm}
\end{figure}

We consider three strategies: \emph{\gls{rx}~$1$} and \emph{\gls{rx}~$2$}, exploiting only the \glspl{cir} collected by each \gls{rx} separately, and \emph{joint \glspl{rx}}, jointly processing the data acquired by both \glspl{rx} in a central entity.
As shown in Fig.~\ref{fig:overall_results}, \gls{rx} $2$ exhibits a mean accuracy 
of $86.59\%$ compared to $84.06\%$ for \gls{rx} $1$. Centralized fusion significantly improves performance, reliability, and stability, increasing mean accuracy to $88.98\%$ while reducing the standard deviation to $0.0319$ compared to $0.0518$ for \gls{rx}~$1$. 
We define the \textit{Danger F1} as the harmonic mean of precision and recall for the critical danger zone, representing the system's ability to correctly identify hazardous scenarios, and the \textit{Safe F1}, which evaluates performance in standard operational zones. Joint processing achieves a mean of $84.94\%$ in danger F1-score, outperforming both \gls{rx} 1 ($80.51\%$) and \gls{rx} 2 ($82.74\%$), and $91.19\%$ in safe F1-score, confirming that the spatial diversity of the multi-\gls{rx} approach enhances classification robustness across all operational states. 

\begin{figure}[t!]
     \centering \includegraphics[width=\linewidth]{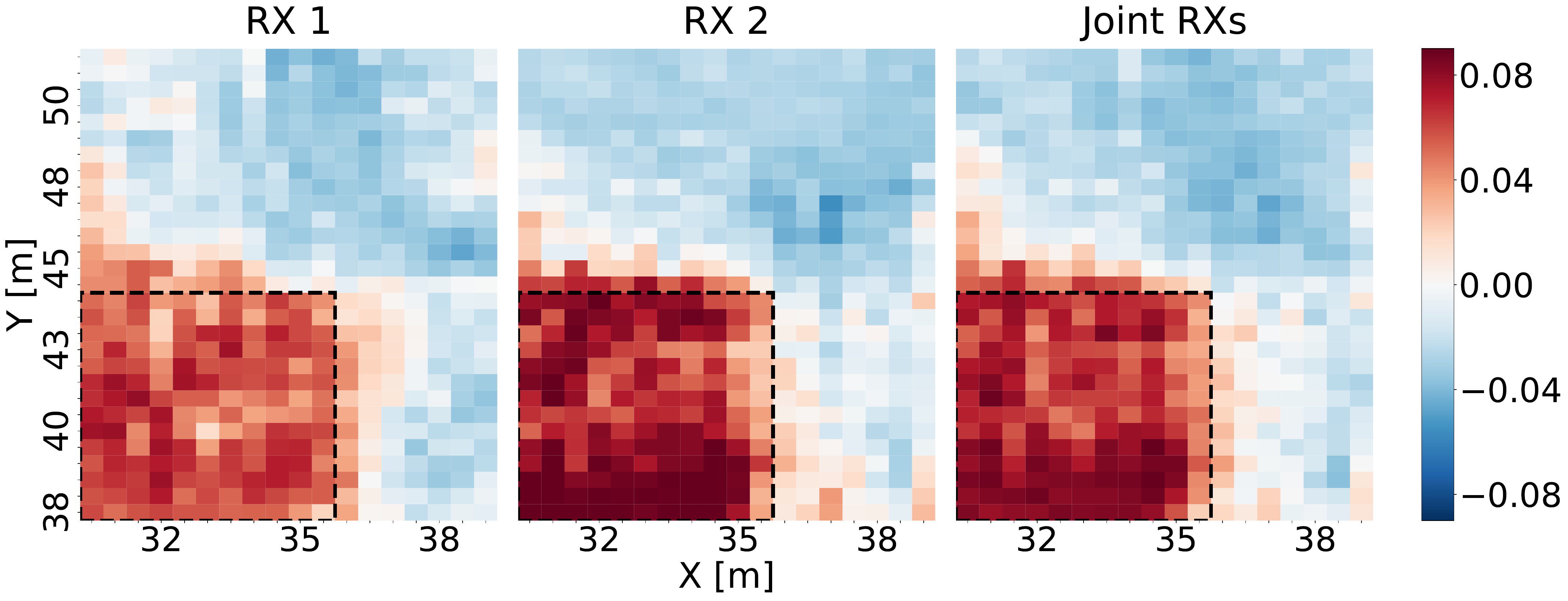}
     \caption{Similarity difference maps (Danger - Safe) for RX 1, RX 2, and Joint RXs. Red/blue areas indicate danger/safe predictions; dashed rectangle delimits the two regions.}
     \label{fig:similarity_diff_heatmaps}
     \vspace{-3mm}
\end{figure}

The spatial bias of the proposed classifier is further investigated through the similarity difference maps presented in Fig.~\ref{fig:similarity_diff_heatmaps}, which illustrate the arithmetic difference between the median similarity scores of the \textit{Danger} and \textit{Safe} classes across the spatial grid. Areas rendered in red indicate a dominance of danger-class similarity, while blue regions signify a stronger safe-class correspondence. A neutral white color represents a decision boundary where similarity scores for both classes are approximately equal. Both \gls{rx} $1$ and \gls{rx} $2$ exhibit a strong and concentrated bias within the designated danger region (indicated by the dashed black rectangle). However, outside this region, particularly in the upper-right quadrant, both individual \glspl{rx} display localized patches of red bias. These \emph{false alarm} gradients suggest that individual sensing streams are susceptible to environmental interference or signal reflections that mimic danger-zone signatures in safe areas. Whereas, by leveraging centralized fusion, the Joint \glspl{rx} configuration mitigates class ambiguity, demonstrating a refined and robust similarity gradient, sharper transition at the borders of the danger region, and uniform blue bias in the safe region.

\subsection{Benchmarking}

In realistic deployments, we envision offline training coupled with real-time inference. Consequently, to accelerate the testing phase, the system infers results by leveraging the binary encoding strategy of the proposed methodology, accepting a marginal trade-off in accuracy for reduced latency. To address this aspect, we compare the results against a baseline resource-heavy \gls{dnn} classifier. We configure the \gls{dnn} following the solution presented in~\cite{benchmarking}, which consists of 8 layers and approximately $465,000$ trainable parameters, optimized via categorical cross-entropy. 

\begin{figure}[t!]
    \centering
    \includegraphics[width=\linewidth]{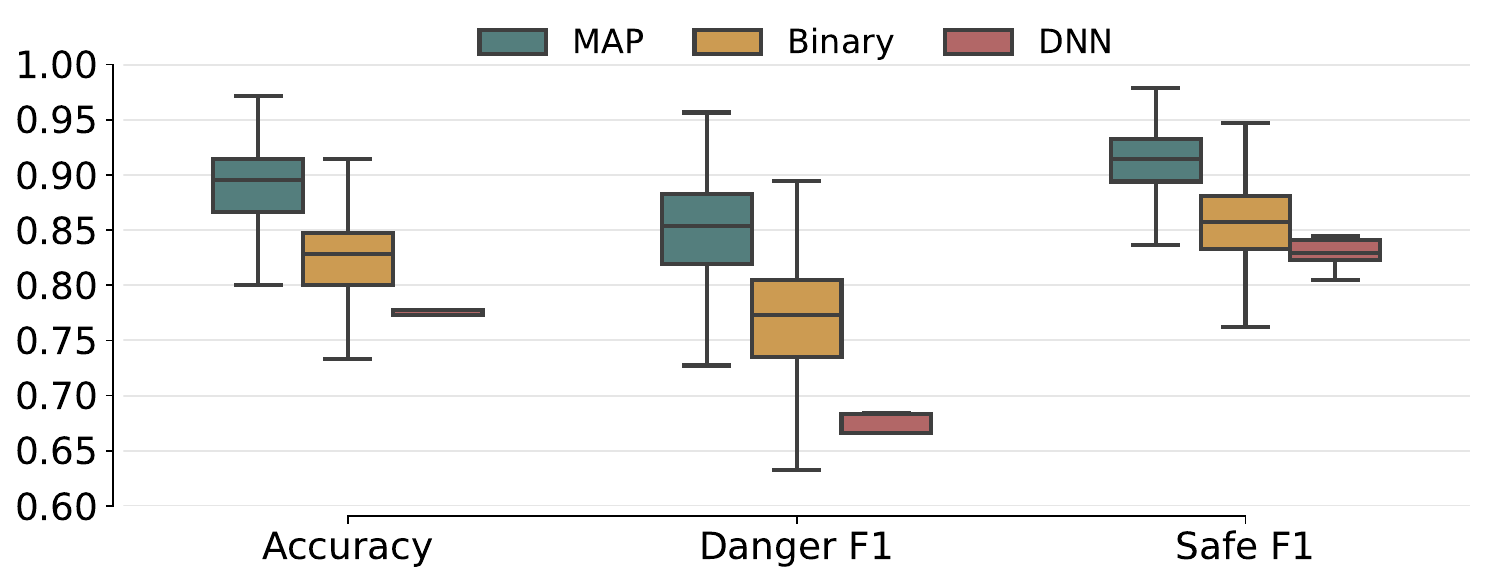}
    \caption{Comparison between the \gls{hdc} encodings and \gls{dnn} baseline in the \emph{joint} scenario.}
    \label{fig:benchmark_results}
    \vspace{-3mm}
\end{figure}

The comparative results illustrated in Fig.~\ref{fig:benchmark_results} highlight that the \gls{hdc}-based approaches, which exploit structured multipath encoding, significantly outperform the \gls{dnn} baseline (operating directly on raw \gls{cir} samples) across all metrics. The continuous \gls{map} encoding achieves the highest performance with a mean accuracy of $88.98\% (\pm 3.19\%)$, surpassing the \gls{dnn} ($77.39\%$) by over $11$ percentage points. The Binary encoding, which is optimized for ultra-fast inference, maintains a substantial lead over the \gls{dnn} with an accuracy of $82.21\%$. Furthermore, the \gls{dnn} struggles to generalize the definition of the Danger zone from the training to the test locations, achieving a Danger F1-score of only $0.6642$, which may be unacceptable in safety-critical applications. In contrast, the \gls{map} approach proves far more robust to spatial variations, achieving a Danger F1-score of $0.8494$. This gap indicates that the highly parameterized \gls{dnn} is prone to overfitting to the specific quantization noise of the training coordinates and demonstrates the advantage of \gls{hdc} in enabling robust, ultra-low-latency inference on resource-constrained edge devices.

\subsection{MCU Performance}
\label{sec:mcu_performance}

To evaluate the feasibility of the proposed \gls{hdc}-based sensing framework in resource-constrained industrial environments, we conducted an execution time analysis on an STM32 Nucleo-L476RG \gls{mcu} featuring an ARM Cortex-M4 core running at $80$~MHz, which is a typical choice for robots and \glspl{agv}~\cite{cobot}. 
The execution time measurements were averaged over $300$ iterations for each fundamental \gls{hdc} operation: bundling, binding, permutation, and similarity check, i.e., \gls{hd} and \gls{cs} for the \gls{map} and Binary approaches, respectively. Standard deviation values are omitted as the variance was found to be negligible across the trials. Table~\ref{tab:hdc_performance} summarizes the results for three different \gls{hv} dimensions $D$. For a standard dimension of $D=10,000$, the binary bundling operation is approximately $17$ times faster than its \gls{map} counterpart ($248$ $\mu s$ vs. $4260$ $\mu s$). Similarly, binding and similarity operations exhibit speedup factors exceeding $15\times$ and $17\times$, respectively.
These results are critical for low-latency industrial safety applications, where the ability to perform high-dimensional permutations and similarity checks in under $200$ $\mu s$ enables the \gls{mcu} to process sensing streams in real-time, leaving sufficient clock cycles for the communication stack. This confirms that binary \gls{hdc} is not only statistically robust, as shown in Section~\ref{sec:NR}, but also exhibits a significant computational advantage for deployment on embedded edge devices. By minimizing the classification overhead, the MCU retains its computational budget for the preceding hardware steps, such as \gls{cir} extraction, facilitating true real-time ISAC operation.

\subsection{ROS Simulation}

As a proof-of-feasibility, we integrate the proposed detection module into the \gls{ros}2-Humble navigation stack to extend and complement environmental awareness beyond traditional onboard sensors such as camera/LiDAR. Specifically, obstacles that are not directly observable within the robot’s onboard sensor \gls{fov} are embedded into the navigation costmap, thereby triggering update of path planning. The proposed module operates as a dedicated \gls{ros} node that estimates the pose and spatial extent of occluded objects in the global map frame. Once an object hypothesis is generated, its region is represented as a 2D point cloud and projected into the original map 
by converting 
it into its occupancy grid. 
Rather than modifying the planner, the system alters the traversability representation by assigning elevated costs to the affected cells. Thus, the global costmap reflects a potential obstacle even though it is not directly observed by the LiDAR, thereby preserving modularity within the \gls{ros} navigation framework and compatibility with standard global planning algorithms. 

\begin{table}[t!]
\centering
\caption{\gls{map} vs. Binary \gls{hdc} execution time ($\mu s$)}
\label{tab:hdc_performance}
\begin{tabular}{l | cc | cc | cc}
\hline
\multirow{2}{*}{\textbf{Operation}} & \multicolumn{2}{c|}{\textbf{D = 100}} & \multicolumn{2}{c|}{\textbf{D = 1,000}} & \multicolumn{2}{c}{\textbf{D = 10,000}} \\ \cline{2-7} 
 & \textbf{MAP} & \textbf{Binary} & \textbf{MAP} & \textbf{Binary} & \textbf{MAP} & \textbf{Binary} \\ \hline
\textbf{Bundle} & 43.0 & 4.3 & 425 & 26.4 & 4260 & 248 \\
\textbf{Bind} & 43.0 & 2.8 & 426 & 14.7 & 4260 & 135 \\
\textbf{Permute} & 9.2 & 3.4 & 76 & 18.3 & 750 & 169 \\
\textbf{Similarity} & 39.0 & 4.4 & 378 & 24.0 & 3740 & 209 \\ \hline
\end{tabular}
\vspace{-2mm}
\end{table}

\begin{figure}[t!]
    \centering
    \includegraphics[width=\linewidth]{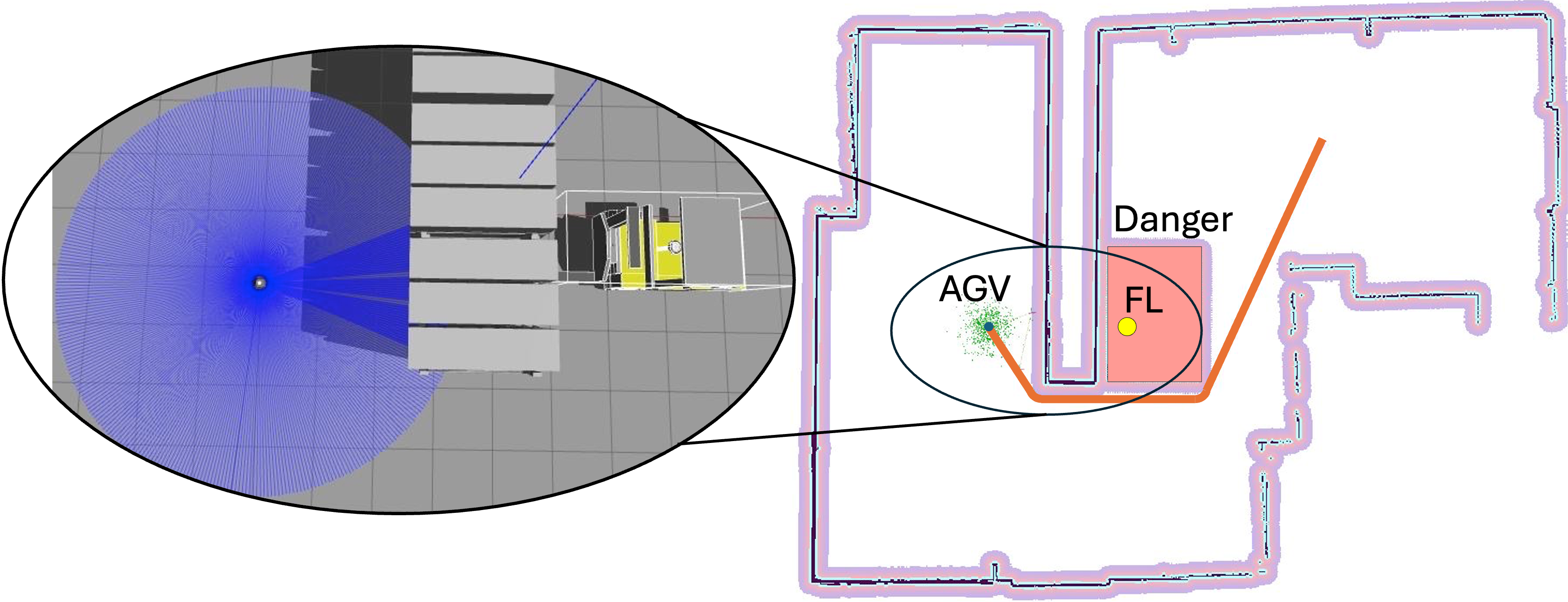}
    \caption{Resulting costmap for a given navigation goal.}
    \vspace{-3mm}
    \label{fig:costmap}
\end{figure}

Fig.~\ref{fig:costmap} illustrates the operational principle. The LiDAR footprint defines the observable region around the robot, while an autonomous \gls{fl} is located behind a shelf and remains outside the direct sensing boundary. The proposed approach infers its presence within the Danger zone, and marks the corresponding costmap area with a high traversal cost. Consequently, the global planner treats this region as non-traversable and generates a path that maintains safe clearance. While the \gls{ros} integration demonstrates improved risk awareness and practical feasibility, we emphasize that the primary benefit lies in embedding risk at the costmap level, enabling earlier awareness in occluded regions, which can trigger conservative behaviors to prevent collisions (such as stopping or re-planning before entering blind spots) rather than relying on reactive corrections. 
This approach offers significant advantages compared to injecting \gls{nlos} information only into the local costmap. Indeed, in the latter case, the robot might initially commit to a trajectory that passes close to the occluded region and only attempt avoidance when approaching it. Such late-stage adjustments may induce risk, especially in human–robot interaction scenarios. In contrast, global cost adaptation ensures that avoidance behavior is embedded in the long-horizon plan, for smoother and more legible motion.

\section{Conclusions}\label{sec:conclusions}

In this paper, we introduced a novel framework that combines distributed devices providing spatial diversity with \gls{hdc}. Unlike conventional onboard sensing systems constrained by \gls{los} operation, the proposed approach leverages infrastructure-based channel measurements and cooperative transmissions to extend perception beyond occlusions in robotic operations. By encoding multipath-induced channel variations into \gls{hv} representations, the system enables robust and scalable inference without explicit channel reconstruction. Numerical results obtained through ray-tracing simulations demonstrate up to $89\%$ detection accuracy in complex indoor environments, while maintaining computational requirements on typical robot hardware. Future work will extend this approach beyond binary classification to support multi-zone risk modeling, enabling finer-grained spatial awareness.

\bibliographystyle{IEEEtran}
\bibliography{refs}

\end{document}